\documentclass{article}

\PassOptionsToPackage{numbers, compress}{natbib}
\usepackage[preprint]{neurips_2025}

\usepackage[utf8]{inputenc} 
\usepackage[T1]{fontenc}    
\usepackage{hyperref}       
\usepackage{url}            
\usepackage{booktabs}       
\usepackage{amsfonts}       
\usepackage{nicefrac}       
\usepackage{microtype}      
\usepackage[table]{xcolor}         
\usepackage{amsmath} 
\usepackage{graphicx}
\usepackage[most]{tcolorbox}
\usepackage{tabularx}
\usepackage{threeparttable}
\usepackage{amssymb}
\usepackage{pifont}
\usepackage{array}
\usepackage{tikz}
\tcbuselibrary{skins}

\title{Simulating the Unseen: Crash Prediction Must Learn from What Did Not Happen}

\author{%
Zihao~Li$^{1}$\thanks{The first two authors contributed equally.}\quad 
Xinyuan~Cao$^{2}$\footnotemark[1] \quad 
Xiangbo~Gao$^{1}$ \quad
\textbf{Kexin~Tian}$^{1}$\quad 
\textbf{Keshu~Wu}$^{1}$ \\
\textbf{Mohammad~Anis}$^{1}$ \quad 
\textbf{Hao~Zhang}$^{1}$ \quad 
\textbf{Keke~Long}$^{3}$ \quad 
\textbf{Jiwan~Jiang}$^{3}$ \quad
\textbf{Xiaopeng~Li}$^{3}$ \\
\textbf{Yunlong~Zhang}$^{1}$ \quad 
\textbf{Tianbao~Yang}$^{1}$ \quad 
\textbf{Dominique~Lord}$^{1}$ \quad 
\textbf{Zhengzhong~Tu}$^{1}$ \quad 
\textbf{Yang~Zhou}$^{1}$\thanks{Corresponding author: Yang~Zhou (\url{yangzhou295@tamu.edu}).} \\
  $^{1}$ Texas A\&M University, $^{2}$ Georgia Tech, $^{3}$ University of Wisconsin-Madison\\
}
\begin{document}

\maketitle

\begin{abstract}
Traffic safety science has long been hindered by a fundamental data paradox: the crashes we most wish to prevent are precisely those events we rarely observe. Existing crash-frequency models and surrogate safety metrics rely heavily on sparse, noisy, and under-reported records, while even sophisticated, high-fidelity simulations undersample the long-tailed situations that trigger catastrophic outcomes such as fatalities. 
We argue that the path to achieving \emph{Vision Zero}, \textit{i.e.}, the complete elimination of traffic fatalities and severe injuries, requires a \textbf{paradigm shift} from traditional \textit{crash-only learning} to a new form of \textit{counterfactual safety learning}: reasoning not only about what happened, but also about the vast set of plausible yet perilous scenarios that could have happened under slightly different circumstances. 
To operationalize this shift, our proposed agenda bridges macro to micro. Guided by crash-rate priors, generative scene engines, diverse driver models, and causal learning, near-miss events are synthesized and explained. A crash-focused digital twin testbed links micro scenes to macro patterns, while a multi-objective validator ensures that simulations maintain statistical realism. This pipeline transforms sparse crash data into rich signals for crash prediction, enabling the stress-testing of vehicles, roads, and policies before deployment. By learning from crashes that almost happened, we can shift traffic safety from reactive forensics to proactive prevention, advancing Vision Zero.

\end{abstract}

\section{Introduction}

Road transportation systems worldwide face a persistent safety challenge, with traffic crashes claiming about 1.35 million lives annually, according to the World Health Organization. In the United States alone, over 40,901 fatal crashes occurred in 2023 \cite{NCSA2025}, imposing a staggering societal cost of estimated \$1.85 trillion, including \$460 billion in direct economic costs and \$1.4 trillion in quality-of-life losses \cite{Blincoe2023}.
Despite decades of investment in road safety research and infrastructure improvements, severe crashes remain stubbornly difficult to predict and prevent, primarily because these catastrophic events are statistically rare relative to the vast scale of driving activity and inherently random. In fact, the US fatality rate stood at just 1.26 deaths per 100 million vehicle-miles traveled (VMT) in 2023 \cite{Blincoe2023}.
This illustrates that even a tiny fraction of driving situations, occurring in mere seconds and meters, can cause tremendous harm, yet pinpointing these events is like finding a needle in a haystack in billions of driving hours~\cite{lord2005poisson}. 
Thus, achieving the ambitious goal of  “Vision Zero”—the elimination of traffic fatalities and serious injuries caused by traffic crashes—demands overcoming a fundamental paradox: the events we most urgently need to learn and avert are those we seldom observe, making traditional reactive approaches fundamentally inadequent~\cite{zhang2025virtual} as it requiring years of driving data to surface then analyze these ``rare events''.
This has underscored the urgent need for innovative predictive frameworks capable of reasoning effectively beyond these statistically rare but devastating traffic scenarios.

\begin{tcolorbox}[colback=blue!10!white,colframe=blue!60!black,title=Rare‑Event Metaphor,width=\linewidth, top=3pt, bottom=3pt]
\small
Imagine a field strewn with a billion identical keys and one hidden landmine that explodes only when its exact key is tried. Thousands of harmless picks give the illusion of safety, yet each trial leaves the catastrophic pairing essentially untested. True safety, then, cannot rely on counting uneventful trials; it requires reasoning explicitly about the \textit{unseen} key‑mine match. So it is with crash prediction: rare, disastrous couplings hide among countless benign moments, demanding insight beyond observed data.
\end{tcolorbox}


To map where and when crashes are most likely, artificial intelligence (AI) has emerged as a promising approach that can proactively mine large crash and exposure datasets with spatial-temporal deep networks.
Researchers have increasingly utilized advanced machine learning methods, including spatio-temporal networks~\cite{yuan2018hetero, zhou2020riskoracle}, Transformers~\cite{karimi2023crashformer,li2025Multimodal}, and diffusion model~\cite{chen2025enhancing}, to learn region-level risk surfaces and patterns that account for traffic flow, road topology, census data, and weather dynamics\cite{wei2024supporting}. Beyond region-level analyses, \emph{surrogate safety measures}, such as time-to-collision (TTC) \cite{HOFFMANN1994511}, conflict counts \cite{cooper1984experience}, post-encroachment time, and other near-miss indicators \cite{HOWLADER2024100331}, enable per-vehicle, second-by-second risk assessment. Existing vision and trajectory-mining algorithms~\cite{suzuki2018anticipating,karim2021system,chen2024evaluating,fang2024abductive,wu2025ai2} automatically extract and label safety-critical signals from video streams and probe-vehicle logs. These metrics are usually embedded in high-fidelity digital-twin simulations. The simulations fuse detailed road geometry, traffic sensors, and weather feeds with learned components, such as GAN-based scenario generators~\cite{deepaccident}, to create virtual testbeds for evaluating vehicle safety functions and traffic interventions under realistic conditions~\cite{zhang2025virtual, kuvsic2023digital,wu2025digital}. Despite these advances, two critical gaps remain: surrogate-based models capture correlations but fail on rare or out-of-distribution events, and simulation pipelines often omit essential real-world details, limiting their fidelity. This paper begins with a mathematical analysis showing how the rarity and randomness of crashes cripple data, hungry AI models. \textbf{We argue that effective “crash AI” must therefore learn from the vastly richer universe of near-misses, events that almost become crashes}. After mapping the strengths and blind spots of current methods, we chart a path to overcome data sparsity, embrace high-dimensional behavioral complexity, and uncover causal insight. The roadmap is built on four mutually reinforcing pillars enabled by suites of AI. Together, these pillars fuse digital twins with generative scenarios, embed multimodal behavior and hypergraph interactions, impose multi-scale causal and symbolic consistency checks, and enable reasoning-driven safety interventions.

\section{Core Challenges in Crash Prediction and Reproduction}
Accurately predicting crashes and reproducing them in experimental settings is intrinsically challenging due to several interrelated factors. We highlight four core challenges that any comprehensive safety analysis framework must address:

\begin{figure}[ht]
    \centering
    \includegraphics[width=0.56\linewidth]{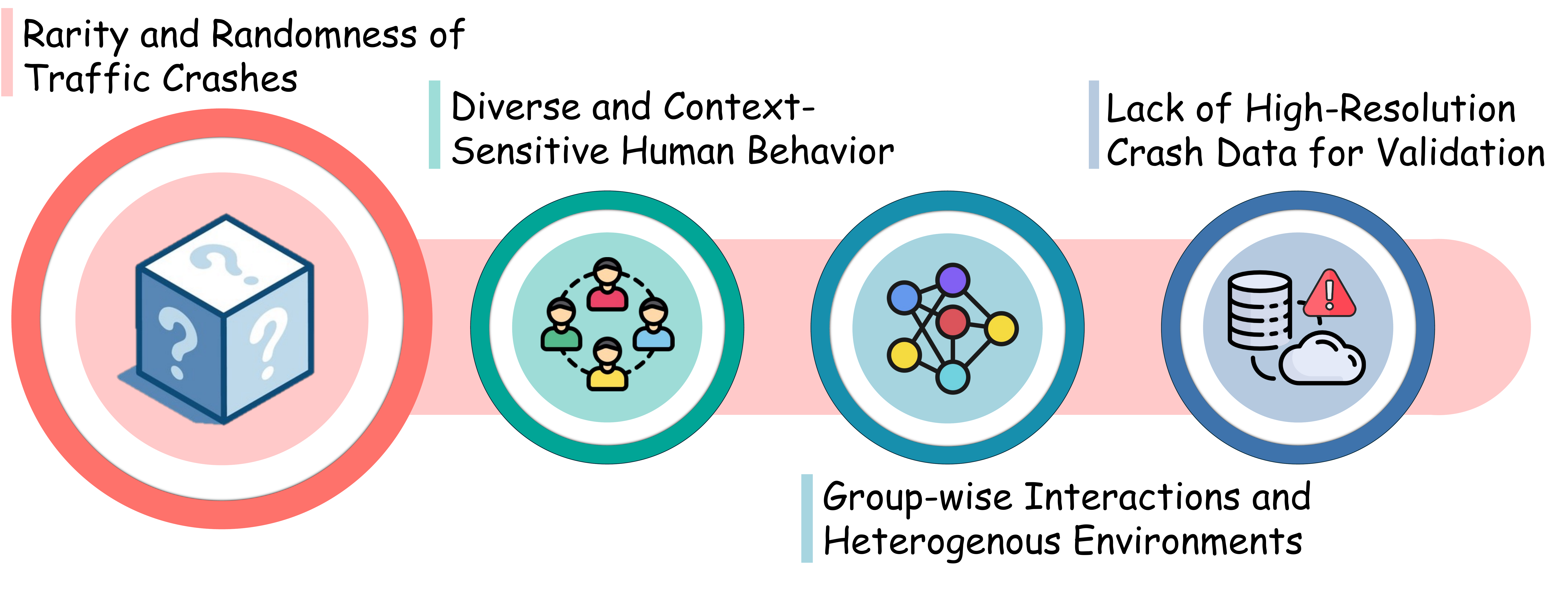}
    \caption{Challenges in Crash-Centric Safety Analysis Framework}
    \label{fig:challenges}
\end{figure}


\subsection{Rarity and Randomness of Traffic Crashes} 

Severe crashes are statistically infrequent and often appear random, which makes it hard to gather sufficient data and validate models \cite{lord2021highway, mannering2020big}.In machine learning terms, rarity corresponds to extreme data imbalance or long-tailed distributions, where crashes make up only a tiny fraction of all driving scenarios. Randomness introduces distributional noise, leading to out-of-distribution edge cases. These factors together challenge standard supervised learning approaches. Traditional crash-frequency analysis requires many years of observations to obtain stable estimates. During this time, vehicle occupants and vulnerable users (e.g., pedestrians, cyclists) get injured, sometimes fatally, and vehicles and different properties or objects get damaged. Even large-scale naturalistic driving studies capture relatively few actual crashes compared to near-misses. For instance, the SHRP2 naturalistic study (with thousands of vehicle-hours of data) recorded on the order of only hundreds of crashes versus thousands of near-crashes. This scarcity means models calibrated on historical crashes risk being either underpowered or overfit to anomalous conditions~\cite{lord2005poisson}. Moreover, the stochastic variability is high – as \cite{hauer2004statistical} noted, crash occurrence has large random fluctuations due to the stochasticity of human nature. Thus, distinguishing meaningful patterns (signal) from randomness (noise) is inherently difficult. Rare-event statistics also challenge model validation: a predictor may perform well on surrogate or aggregated metrics yet still fail to capture truly rare extreme cases. Although AUC maximization~\cite{yuan2021compositional,yang2021deep,yuan2021large} offers a promising approach to address class imbalance in such settings, AUC-based objectives remain difficult to optimize under significant data randomness and out-of-distribution conditions~\cite{yang2022auc}. 

In summary, crash data are “small data” in a big data world – any reliable prediction strategy must somehow amplify information from non-crash surrogates or simulations to overcome the rarity of the events of ultimate interest. To our knowledge, this is the \textbf{first systematic mathematical analysis of safety learning that explicitly accounts for both randomness and rarity}; to further illustrate the inherent difficulty of learning from such rare, stochastic events, we present a simplified mathematical formulation using Fisher information~\cite{fisher1925theory} to quantify the limits of estimating crash probabilities from crash-only data (Appendix~\ref{proof:fish_rarity}-\ref{proof:fish_stochastic}). As a result, rare-event estimators remain highly sensitive to sampling noise, limiting reliability. \textbf{Crash-only learning is therefore statistically inefficient}. To address this, we advocate for a shift toward counterfactual reasoning (Appendix~\ref{proof:aug}). Many non-crash scenes exhibit high-risk behaviors, such as unsafe following or delayed reactions, that reflect elevated latent risk. We propose augmenting the dataset with near-miss cases where the model-estimated crash probability exceeds a threshold. \textbf{High-risk near misses densify informative samples, sharpen decision boundaries, boost Fisher information, and lower estimator variance, enabling efficient learning without extra crash data}.

\begin{tcolorbox}[colback=blue!10!white,colframe=blue!60!black,title=Hungry Models,width=\linewidth, top=3pt, bottom=3pt]
\small
Crash-only learning starves on rare events, and randomness only deepens its hunger.
\end{tcolorbox}

\subsection{Diverse and Context-Sensitive Human Behavior} 
Besides the aforementioned curse of rarity~\cite{lord2005poisson,liu2024curse}, human drivers exhibit extremely complex and varied behavior, which becomes even less predictable in emergency or time-critical situations. Driving decisions are diverse\cite{kosaraju2019socialbigat, gao2020socialstgcnn, zhora2023what, bhat2020multimodal, mcallister2020multimodal} – a driver may choose to brake, swerve, or accelerate in response to a hazard, and this choice depends on a host of factors (individual skill, reaction time, attention state, surrounding traffic, road condition, etc.) as suggested by Yerkes-Dodson Law \cite{yerkes1908relation}. Critically, under emergency imminent-crash scenarios, human reactions can be highly context-sensitive and sometimes suboptimal. For example, some drivers instinctively brake hard while others may swerve or do nothing (freezing up) when a sudden obstacle appears \cite{wang2016driversPartC}. Such differences in human response can tip the scale between a collision and a narrow escape. Yet capturing these nuances in models is challenging. Simple rules or distributions may not reflect how humans behave in rare panic situations. In essence, human-in-the-loop uncertainty is a major hurdle: realistic safety analysis needs to model not just an average driver but the whole spectrum of possible driver behaviors and errors, especially in critical moments \cite{singi2024decision, shao2024lmdrive,qian2023end}.

\begin{tcolorbox}[colback=blue!10!white,colframe=blue!60!black,title=Stochastic Human Nature,width=\linewidth, top=3pt, bottom=3pt]
\small
Human unpredictability is the hurdle, true safety comes from modeling every behavior, not just the average one.
\end{tcolorbox}


\subsection{Group-wise Interactions and Heterogeneous Environments} 
Traffic crashes frequently result from the collective dynamics of multiple road users, vehicles, pedestrians, and cyclists, navigating complex and variable environments \cite{sahu2025physics,zernetsch2024detecting}. Group‐wise interactions, defined as the dynamic interplay among numerous actors whose behaviors mutually influence one another over time, occur both within each user class (e.g., vehicles responding to other vehicles \cite{wu2025ai2,wu2023graph,wu2024hypergraph}; pedestrians adjusting to fellow pedestrians \cite{trumpp2022modeling}) as well as across classes (e.g., vehicle–infrastructure exchanges \cite{colley2024pedsumo} or mixed vehicle–pedestrian flows \cite{zhang2022learning}

\textbf{In transportation engineering, “human–vehicle–environment” interaction is the foundational concept taught in the very first course} \cite{manual2000highway}, \textbf{yet traditional model‐based approaches often fail short in capturing this high‐dimensional complexity} \cite{wu2025ai2}. For example, a single lane‐change maneuver in dense traffic can trigger oscillatory shockwaves: as one vehicle shifts lanes, following vehicles brake and accelerate in response, generating stop‐and‐go waves that propagate upstream and amplify risk \cite{shi2022integrated,li2024enhancing, shi2025predictive, zhang2025anticipatory, tian2025physically}. Furthermore, in dense traffic, a sudden deceleration by one vehicle within a tightly packed convoy may precipitate a multi‐vehicle collision through rapid propagation of traffic waves \cite{sugiyama2012multiple,salzmann2020trajectron++}. Roadway characteristics, including lane geometry, intersection control mechanisms, sight‐distance limitations, and surface conditions, vary substantially between urban, suburban, and rural contexts, further shaping how these group interactions evolve \cite{yuan2018hetero}. Macro‐scale context therefore matters: an emergency stop on a congested freeway differs fundamentally from one on an empty rural road \cite{liao2024crash}. Effective crash prediction must be context‐aware, capturing how risk emerges from the interaction of many agents under specific environmental conditions \cite{wang2024multimodal,grigorev2023traffic}.

\begin{figure}[ht]
    \centering
    \includegraphics[width=0.8\linewidth]{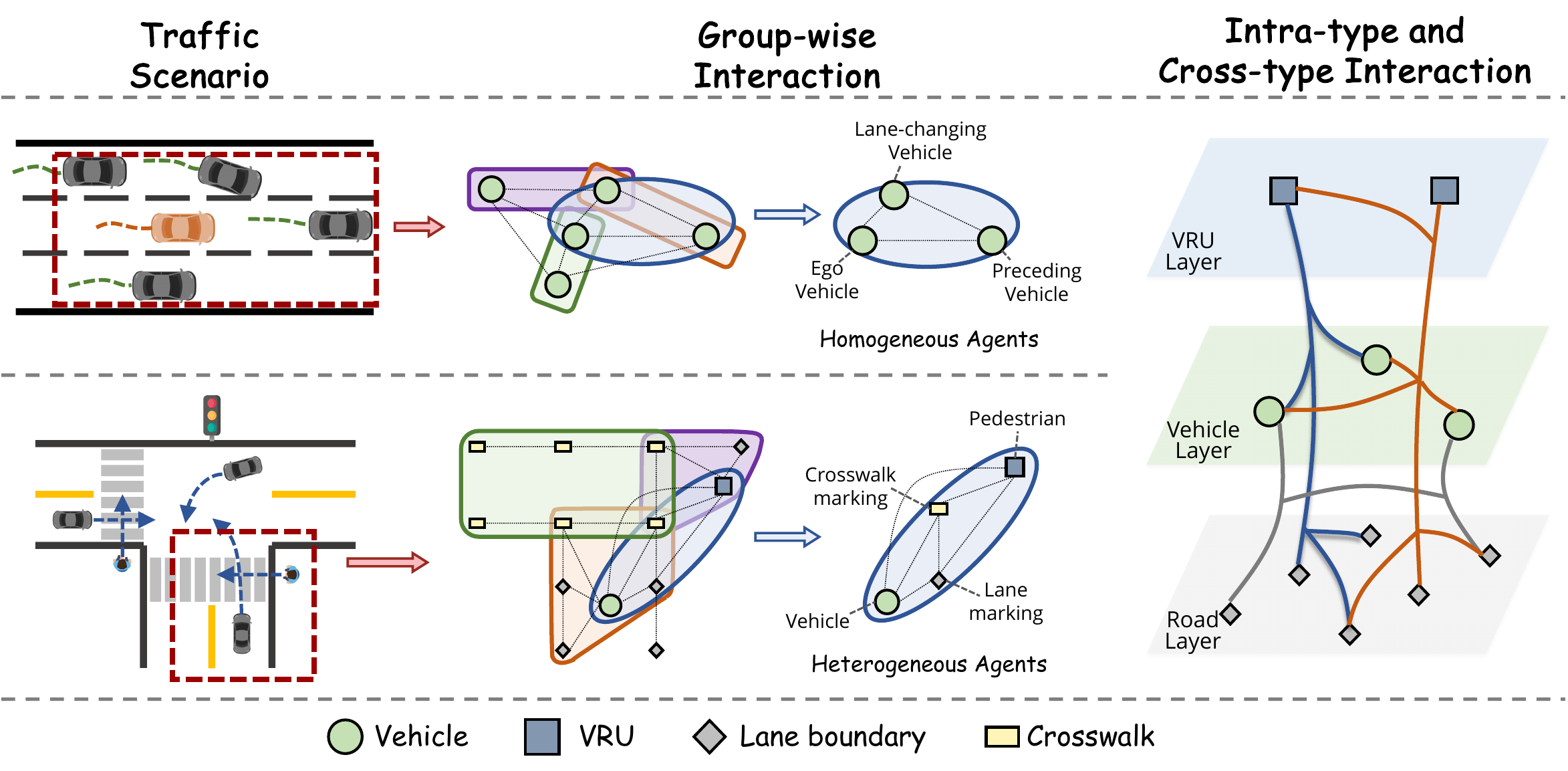}
    \caption{Context‐aware group‐wise interactions and both intra‐type and cross‐type relations in traffic, illustrating higher‐order graph dynamics dependencies.}
    \label{fig:interaction}
\end{figure}

\begin{tcolorbox}[colback=blue!10!white,colframe=blue!60!black,title=Emergent Multi‐Agent Risk,width=\linewidth,, top=3pt, bottom=3pt]
\small
Traffic risk emerges from drivers interacting with their specific environmental context and, when other road users are present, from interactions within that context, underscoring the need for context-aware, group-wise, multi-agent modeling.
\end{tcolorbox}


\subsection{Lack of High-Resolution Crash Data for Validation} 
Macroscopic crash databases (Table~\ref{tab:trad_crash_datasets}), including state and national compilations of police reports and roadway inventories, offer millions of crash records describing basic information such as location and time. Yet these sources are inherently coarse and often under-reported. Moreover, they typically lack contextual information. Microscale resources, by contrast, stream high-rate, multi-modal sensor data that precisely track vehicle trajectories, driver inputs, and environmental context (Tables~\ref{tab:NDS}-\ref{tab:traffic_datasets}).  However, serious crashes are rare in these collections, leaving models to train on plentiful surrogate or near-miss events while lacking ground-truth crash dynamics. \textbf{We know how people drive, but not how they crash within seconds}.
Even the context‑rich SHRP‑2 study \cite{Shrp2} captured only~1–2k mostly minor crashes across millions of miles, with many events missing synchronized multi‑modal data (Table~\ref{tab:NDS}). Crash datasets are wide captured but shallow, with millions of events with minimal detail. Without richer multi-modal crash records to support AI reasoning, it remains difficult to validate that simulations or analytical models accurately reproduce real-world impacts.

\begin{tcolorbox}[colback=blue!10!white,colframe=blue!60!black,title=Summary Challenges,width=\linewidth, top=3pt, bottom=3pt]
\small
Crash rarity, behavioral complexity, multi‑agent dynamics, and data scarcity constrain the prevailing safety analysis. AI methods that fuse heterogeneous real and simulated data offer great potential.
\end{tcolorbox}

\section{Limitations of Current Methods}
While recent studies have advanced both macro- and micro-level crash modeling, limitations persist at each scale. We begin by examining macroscopic crash-frequency models, then turn to micro-level approaches such as surrogate safety metrics and simulation. We conclude by highlighting the disconnect between these levels, which hinders a unified understanding of traffic safety.

\subsection{Limitations of Macroscopic Crash‐Frequency Learning}

Over the past decade, AI-based crash-frequency research has reframed safety prediction as a spatio-temporal learning problem on road graphs. Convolutional and graph-neural architectures ingest high-resolution traffic probes, weather data, and points of interest, and output fine-grained risk maps~\cite{yuan2018hetero,zhou2020riskoracle,nippani2023graph}. Yet they learn purely correlational patterns from police reports that underreport 30–40\% of minor or non-injury crashes~\cite{janstrup2016understanding}, and from crash-count datasets dominated by zeros due to event rarity over short time frames or road segments~\cite{lord2018safety,lord2018safe}. Table~\ref{tab:trad_crash_datasets} summarizes the traditional crash datasets. 

Macroscopic crash-frequency statistical modeling remains rooted in count-data formulations, such as Poisson, negative-binomial, Poisson–lognormal, negative binomial-Lindley, and their spatio-temporal or random-parameters extensions \cite{lord2010statistical,lord2021highway}. While these models yield interpretable estimates and support more defensible causal inference, they struggle to scale to massive, high-dimensional road–time grids due to computational bottlenecks, and their linear link functions cannot natively capture complex, multimodal interactions without extensive, manual feature engineering \cite{lord2021highway, mannering2020big}. To bridge this trade‐off, Hybrid frameworks merge statistical models with deep learning~\cite{li2025Multimodal,dong2018improved,jin2023real}, gaining the non-linear flexibility absent from classical approaches and offering a clearer path toward causal inference. Yet, because they still rely on aggregated crash counts, the insights remain macro-level and often do not generalize to specific locations or events, limiting their usefulness for targeted safety interventions.

 \begin{tcolorbox}[colback=blue!10!white,colframe=blue!60!black,title=Correlation Is Not Enough,width=\linewidth, top=3pt, bottom=3pt]
\small
 Correlation is useful for prioritization, but causality is indispensable for intervention design.
\end{tcolorbox}

\subsection{Reliance on Surrogate Safety Metrics with Uncertain Crash Correlation} 
Surrogate safety measures (SSMs), and other “near-miss” indices, provide an anticipatory view of crash risk \cite{wang2021review}. Typically, SSMs are computed by solving ordinary differential equations (ODEs) that model vehicle kinematics and flag an event whenever inter-vehicle separation falls below physical dimensions \cite{zhang2025anticipatory, li2024beyond,li2024disturbances}. These metrics excel at highlighting hazardous interactions that occur far more often than actual crashes. \textbf{Yet mapping moderately low TTC values (e.g.\ 1.0–2.5 s) to true collision risk remains ambiguous: unless TTC falls into a very low regime, there is no consensus on what threshold signifies danger}. The jump from a moderately low SSM value to an actual collision is tenuous, as countless sub-1.5 s TTC events resolve safely. This occurs because SSMs depend critically on the chosen vehicle‐dynamics model, assumed driver‐behavior parameters, and encoded interaction scenarios. Consequently, correlations between SSM counts and crash counts fluctuate with context, threshold choice, and study design, making SSMs reliable for ranking relative severity but unreliable for quantifying absolute risk. Although Extreme Value Theory has been used within hierarchical Bayesian frameworks to extrapolate crash likelihoods from SSM tails \cite{tarko2018estimating,zheng2014freeway,anis2025real}, it still relies on strong, hard-to-verify assumptions about tail shape and does not systematically resolve these fundamental ambiguities.

\begin{tcolorbox}[colback=blue!10!white,colframe=blue!60!black,title=SSMs may not work,width=\linewidth, top=3pt, bottom=3pt]
\small
Put simply, SSMs only earn our trust as they near zero. At that point they’re critical alarms, but at moderately low values their indication of true crash likelihood remains murky.
\end{tcolorbox}


\subsection{High-fidelity simulation still under-samples the crash tail}

State-of-the-art simulators, combining HD maps, multi-body dynamics, and sensor emulation, reproduce everyday traffic with impressive detail. The details of the existing simulation are summarized in Tables~\ref{tab:simtool_overview}-\ref{tab:llm_simtool_ow} of the Appendix. Extensions such as stochastic driver models~\cite{li2024disturbances,albaba2019stochastic,li2025adaptive}, domain-randomized physics~\cite{amini2020learning}, reinforcement-learning agents~\cite{suo2021trafficsim, li2022metadrive,yang2023improved}, and adversarial~\cite{li2019aads,feng2021intelligent,yan2023learning} or importance-sampling searches~\cite{feng2023dense, koren2018adaptive}  have improved fidelity for typical maneuvers. Yet critical edge cases remain under-represented. Driver models, even when randomly perturbed, fail to capture the full spectrum of human error, and physical parameters are almost always randomized independently. As a result, \textbf{some dangerous joint conditions (e.g., low tire–road friction + delayed braking + poor visibility), which may lie well outside any single marginal tail, occur neither realistically nor with controllable frequency}. While marginal long-tail events in any one parameter (such as extremely low friction or severe sensor noise) are themselves high-risk, validation typically measures only aggregate exposure (total simulated kilometers) rather than explicit coverage of these high-risk combinations, offering no guarantee that the rare-event manifold has been explored. Consequently, \textbf{simulators can score well on overall statistics yet still miss the critical collision modes that drive real-world fatalities, fostering unwarranted confidence in their safety assessments}.

\subsection{Macro-micro inconsistency} A persistent limitation in current traffic safety research is the disconnect between macro-level and micro-level analyses~\cite{cai2019integrating}. Road safety management often emphasizes system-wide indicators, such as crash rates per million vehicle-miles or the identification of high-risk locations (black spots), but overlooks the fine-grained dynamics that cause individual crashes \cite{nippani2023graph, chen2016learning,krumm2017risk,najjar2017combining}. Conversely, micro-level studies, such as simulations or driving simulator experiments~\cite{suzuki2018anticipating, fang2024abductive, suo2021trafficsim, bergamini2021simnet, montali2023waymo}, typically examine specific behaviors or events in isolation, without accounting for their broader implications on network-level safety. Furthermore, many simulation platforms are primarily designed for analyzing traffic efficiency and do not explicitly model collisions or safety-critical outcomes. This disconnect between purpose and application introduces inconsistencies: micro-level insights may fail to generalize or translate to measurable improvements in real-world safety, especially when the tested conditions are rare or the operational environment differs significantly, a phenomenon known as distribution shift~\cite{suo2021trafficsim, yan2023learning, stoler2024safeshift}. Without a mechanism to bridge these scales, we lack the ability to verify whether micro-level innovations produce system-level benefits, or to prioritize which micro-level scenarios matter most based on macro-level crash data.\textbf{Closing this loop requires a unified, bidirectional framework} in which macro crash patterns inform the generation of critical micro-level scenarios, and micro-level causal evidence is fed back to refine and calibrate macro-level risk models.

\begin{tcolorbox}[colback=blue!10!white,colframe=blue!60!black,title=Macro–Micro Divide,width=\linewidth, top=3pt, bottom=3pt]
\small
Accurate micro-level behavior models do not ensure macro-level safety gains. Without integrating insights across scales, detailed fidelity stays inconsequential.
\end{tcolorbox}


\section{Toward Vision Zero: An AI-Centered Agenda}
To systematically overcome the core challenges inherent in traditional methods, we propose an integrated AI-driven framework structured around four foundational pillars: (1) New Simulation Platform (a high-fidelity digital twin for perturbation-driven safety evaluation), (2) New Scenario Engine (an advanced generative AI system for realistic and rare-event scenario synthesis), (3) New Validation Suite (multi-scale robustness metrics and rare-event validation), and (4) New Intervention Platform (actionable risk insights powered by interpretable AI reasoning and reinforcement learning).
This unified approach explicitly quantifies and identifies true crash vulnerabilities, transitioning from ambiguous surrogate measures to direct, scenario-specific risk quantification through systematic perturbation, generation, validation, and intervention.

\begin{figure}[ht]
    \centering
    \includegraphics[width=0.8\linewidth]{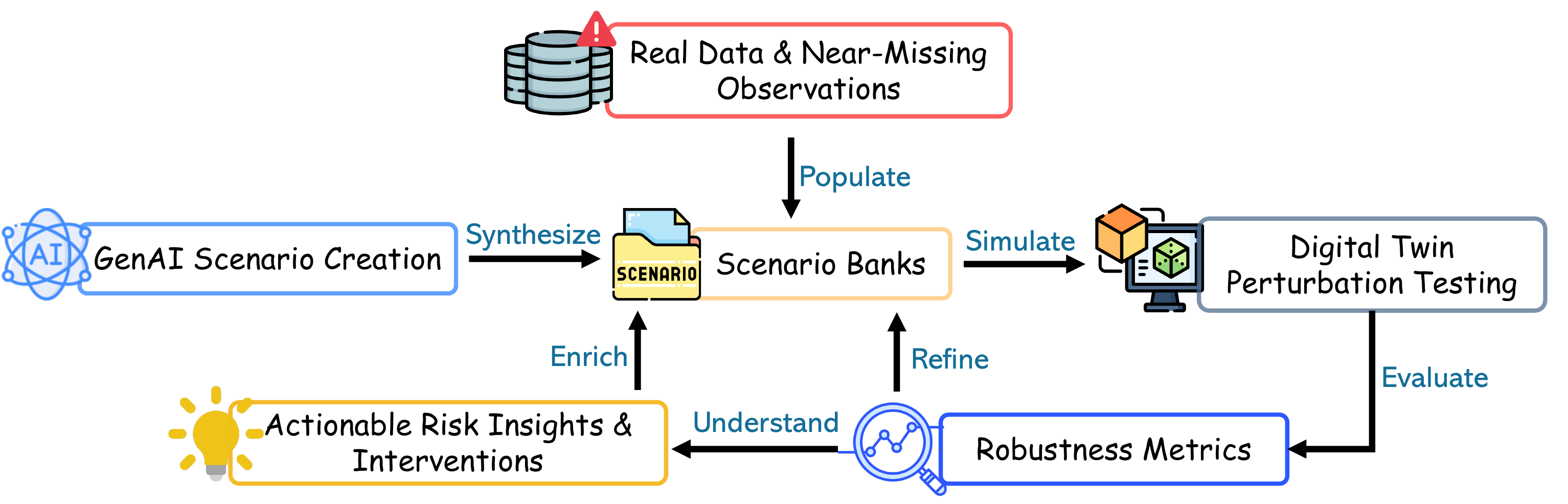}
    \caption{A pipeline integrating GenAI-driven scenario creation, digital twin perturbation testing, and robustness evaluation to uncover rare-event risks and support Vision Zero interventions.}
    \label{fig:net-trans}
\end{figure}

\subsection{Generative AI Techniques for Realistic Scenario Creation and Behavior Emulation}

\paragraph{Generative AI for Scenario Creation.}
Recent progress in deep generative modeling offers a data-driven alternative to hand-crafted test suites. By learning directly from large-scale real-world driving datasets, modern models can reproduce the entire spectrum of traffic states in multiple formats, including time-series trajectories, rasterized semantic maps, video-like image sequences, and simulated sensor streams. The generated “AI worlds" jointly sample \emph{agents} (vehicle positions, velocities, goals), \emph{environment} (road geometry, signage, occlusions), and \emph{conditions} (lighting, weather, work zones), providing variability that manual scripting struggles to reach.Current research follows three converging threads. First, utilizing reinforcement learning with a reward function consisting of both plausibility objectives \cite{liu2024safety}. Second, conditional generative models, ranging from GANs to diffusion processes, use structured priors such as lane topology~\cite{rowe2025scenario}, initial traffic layouts~\cite{NEURIPS2024_64ff8d0b}, or free-form natural-language prompts~\cite{tan2023language,zhang2024chatscene} to yield controllable, diversified scenarios. Third, large transformer-based diffusion models scale this idea further, capturing macro traffic context and micro interaction details in a single pass and allowing fine-grained scene editing or interpolation~\cite{peebles2023scalable,guo2023controllable}. Together, these advances recast scenario generation as principled sampling over rich spatio-temporal and sensory data spaces, supplying diverse, realistic, and tunable test cases for both simulation and on-road evaluation.

\paragraph{Personalized and Multi-modal Driver Behavior through AI.}
Generative AI now supports driver models spanning the full spectrum of human behaviors rather than a single rule set. By training on large naturalistic-trajectory datasets, these models learn latent embeddings that capture diverse behaviors, such as cautious, aggressive, distracted, and highly responsive~\cite{zhang2023proactive}. Sampling from these embeddings in simulation populates scenes with heterogeneous agents, enhancing realism and behavioral diversity~\cite{shao2024lmdrive,qian2023end}. Variational autoencoders encode continuous intention manifolds, while GAN/GAIL variants emphasize rare, high-risk maneuvers~\cite{mescheder2017adversarial,roy2019vehicle}. As a result, AI drivers exhibit varied reaction-time distributions, gap-acceptance thresholds, and lane-change propensities, escalating to panic responses under stressors such as phantom braking or adversarial disturbances~\cite{tee2023learning,gao2024collision,nassi2020phantom}. This fusion of generative realism with adversarial focus yields traffic streams that stress-test both everyday flow and low-probability hazards.

\paragraph{Hypergraph‐based modeling of interactive environments.}
Graph neural networks (GNNs) excel at encoding \emph{pair-wise} agent dependencies, yet real traffic risk emerges from \emph{group-wise} interactions involving vehicles, VRUs, and infrastructure \cite{battaglia2018relational}. Hypergraphs generalise GNNs by allowing one edge to bind multiple heterogeneous nodes, thereby capturing collective manoeuvres such as platoon oscillations or pedestrian–vehicle negotiation at crossings \cite{wu2024hypergraph,wang2025nest}. To represent these rich relations faithfully, a \emph{heterogeneous-structured, temporally evolving hypergraph} is needed: nodes of different types (vehicle, pedestrian, traffic control devices, roadway cell) participate in hyperedges whose composition and strength change over time, learned via dynamic attention mechanisms \cite{zhang2025learning,ruggeri2024framework,hayat2024heterogeneous}. This potential structure lets the model reason over high-order dependencies while adapting online to the ever-shifting topology of real traffic streams.

\paragraph{Generative AI for Crash and Near-Crash Events.} 

One core promise of generative modeling lies in its ability to overcome the rarity of crash data by creating synthetic yet plausible near-crash or crash scenarios. These outputs can take the form of time-stamped sequences of agent states, simulated video frames, or multimodal logs consistent with real crash precursors. Current approaches include class resampling to rebalance rare outcomes \cite{o2018scalable}, adversarial perturbation of real-world scenes \cite{koren2018adaptive, rempe2022generating, mei2025seeking}, diffusion-based generative modeling for rare events \cite{chen2025risk, xie2024advdiffuser,xu2025diffscene}, and reinforcement learning with failure-driven objectives \cite{kulkarni2024crash}. To maintain the realism and statistical validity of generated data, post-processing steps like rejection sampling are employed to align the synthetic data with real-world crash distributions, for example, by matching empirical annual crash counts per vehicle-kilometer \cite{feng2023dense}. This step ensures that the overall datasets remain statistically consistent with macroscopic safety data.

\begin{figure}[ht]
    \centering
    \includegraphics[width=1\linewidth]{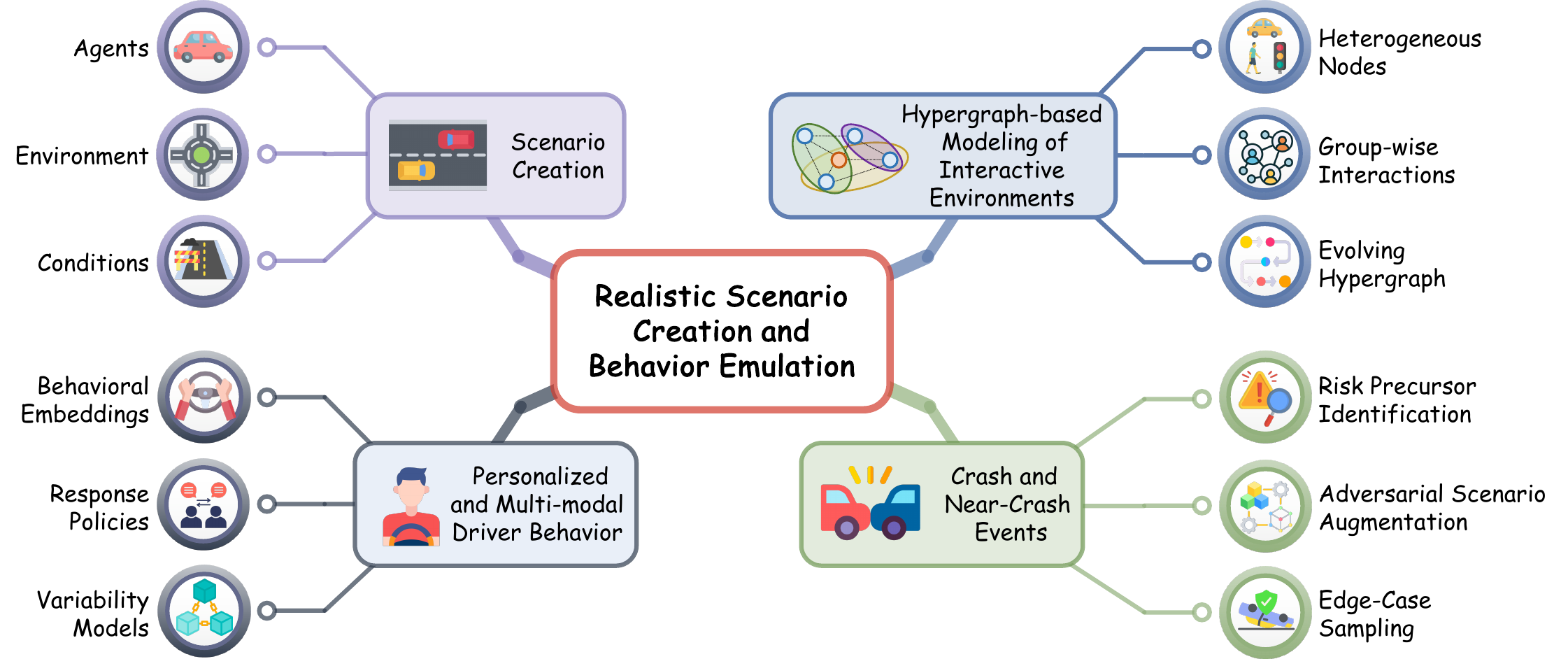}
    \caption{Mind-map of generative AI techniques for scenario creation and behavior emulation.}
    \label{fig:scenario_behavior_mindmap}
\end{figure}

\subsection{New Platform: AI-Driven Digital Twin and Human-in-the-Loop (HITL) Design}

Leveraging the AI techniques outlined above, our proposed high-fidelity digital twin directly addresses crash rarity, behavioral unpredictability, and multi-agent interaction. Built on PhysX~\cite{NVIDIA_PhysX_5_6_0} and CARLA~\cite{dosovitskiy2017carla}, the platform fuses precise road geometries, high-order vehicle dynamics, and realistic environmental factors (e.g., weather, visibility, and infrastructure states) into a single unified virtual environment~\cite{zhang2025virtual}.

Beyond the digital twin mentioned above, traffic simulators integrated with physical driving simulators also contribute to further examining the behaviors, which require photorealism. The quality of visual representation directly impacts the validity of human behavioral responses within simulated environments.
 Recent neural-based 3D reconstruction methods offer efficient ways. Neural Radiance Fields~\cite{mildenhall2021nerf, he2024neural} and Gaussian Splatting~\cite{kerbl20233d, yan2024street} enable the creation of visually convincing digital replicas from captured images. Although these methods excel at generating photorealistic visual representations, they often lack the geometric information needed to simulate physical interactions such as collisions and surface friction. These elements are essential for accurate traffic simulations.
To bridge this gap between visual fidelity and physical plausibility, geometric-aware reconstruction algorithms have emerged as promising solutions~\cite{huang20242d, guedon2024sugar, chen2023neusg}, facilitating more physically accurate simulations while maintaining visual quality. 
For further enhancing the visual fidelity of digital assets, advanced generative AI models can effectively predict and fill the residual gaps between simulation and reality~\cite{alhaija2025cosmos, fan2024freesim}, producing more coherent and complete representations.

\subsection{New Suites of Evaluation and Reasoning : Multi-Scale Evaluation and Causal Reasoning}
The new framework must also rethink how to validate and evaluate our safety analysis methods. Instead of relying on traditional validation that focuses on a single domain, we propose a suite of multi-scale hybrid validation metrics that combine indicators of micro-level realism \cite{song2020pip} with macro-level safety outcomes \cite{yan2008validating}. This ensures micro‐level realism and enables two additional metrics: (1) Crash surrogate alignment, measuring how well surrogate‐identified high‐risk areas match historical crash hotspots; and (2) Rare-event reproduction index: the statistical similarity between the distribution of outcomes from simulations seeded with real-crash scenarios and the distribution of observed crash events. A multi‐objective Pareto search balances fine‐grained simulation fidelity, surrogate alignment, and rare‐event reproduction, stopping when no further Pareto gain is possible~\cite{veran2023Pareto,marler2004survey}.

\paragraph{Causal reasoning.}
While the aforementioned metrics provide direct quantitative measurement, they lack interpretability and the ability to enforce known constraints, both of which are especially important in scenarios involving crash prediction. In this sense, causal reasoning can be further incorporated to corroborate the understanding with AI engines: 
(i) \emph{Structural causal models (SCMs).} An SCM is a directed acyclic graph whose nodes are traffic variables (speed, friction, gap, \emph{etc.}) and whose structural equations define how interventions propagate; do-calculus enables counterfactual queries, e.g., “Had friction been higher, would the crash persist?”~\cite{neuberg2003causality,bareinboim2016causal,zhu2019causal};  
(ii) \emph{SCM-conditioned LLMs} generate counterfactual ``what if'' scenario reconstructions, making such counterfactual queries scalable to high-dimensional multimodal traffic states. This pairs each explanation with natural-language rationales generated by a Vision Language Model (VLM) fine-tuned on “traffic-narration” data, allowing stakeholders to inspect causal chains without deciphering raw tensors.~\cite{gendron2024counterfactual,wei2022chain,pawlowski2020deep};  
(iii) \emph{Symbolic and Neuro-symbolic validators.} Symbolic AI encodes high-level traffic rules, such as right-of-way laws, reaction-time limits, and common failure modes, using formal rules~\cite{sarker2022neuro}. These rules act as scenario filters, or merge with neural modules via differentiable logic layers to create hybrid neuro-symbolic checks~\cite{manhaeve2018deepproblog,hitzler2022neuro,garcez2019neural}. Embedding structured reasoning in this way boosts both robustness and interpretability of the system.

\subsection{New Intervention Brain: Intervention Design with Reasoning}
Building upon the risk insights and causal validation obtained from the multi-scale evaluation suite, the final step is to design intelligent, reasoning-driven interventions that mitigate identified crash scenarios. At the core of our intervention platform lies Reinforcement Learning (RL), which frames intervention design as an optimization problem: minimizing the probability of crashes under high-dimensional, uncertain conditions. The digital twin environment (Sects. 4.1–4.2) and rare-event reward shaping (Sect. 4.3) naturally support this formulation, making RL a foundational component for reasoning-driven decision making.
Within this high-fidelity simulator integrated with AI components, RL agents explore and refine control policies by interacting with a wide spectrum of crash-prone scenarios. Variants such as adversarial RL~\cite{pinto2017robust, ilahi2021challenges}, robust RL~\cite{morimoto2005robust}, and hierarchical/hybrid RL~\cite{pateria2021hierarchical, yue2024hybrid} offer specialized mechanisms for handling uncertainties and maintaining performance under worst-case disturbances. However, conventional RL lacks semantic grounding, interpretability \cite{glanois2024survey}, and generalization to previously unseen rare events. To address these limitations,  reasoning-enhanced extensions are needed, that augment RL with VLMs. VLM-augmented RL integrates high-level semantic cues and prompt-based guidance, using retrieved memory from historical scenarios to inform exploration and action selection \cite{long2024vlm, sheng2025curricuvlm}. The VLM supports RL by offering interpretable rationales \cite{huang2024vlmrl}, context-aware priors \cite{li2024dialogue}, and natural language feedback during training \cite{duan2024prompting} to improve policy stability and correct unsafe behavior. Beyond the RL paradigm, Vision–Language Action (VLA) ~\cite{brohan2023rt, kim2024openvla, black2024pi0visionlanguageactionflowmodel} providing an interpretable alternative by mapping perceptual inputs directly to intervention actions through pretrained models is also a viable solution. The effort in Sections 4.1 to 4.4 will form a unified framework that advances our mission toward Vision Zero by transforming identified risk into intelligent and actionable policy.

\section{Conclusions}
Crash prediction will remain inconsistent and data‑starved until we couple network‑level statistics with trajectory‑level details and augment both with plausible counterfactual scenarios. In this position paper, we argue that counterfactual augmentation, causal learning, and chain-of-thought reasoning with multimodal large models offer a path forward. Generative “what-if” engines simulate unobserved crash scenarios, causal graphs reveal why a near-miss becomes a crash or remains safe, and large language models translate these causal pathways into clear, human-readable explanations. Together, they transform sparse crash records into rich training signals for crash prediction. Implementing this pipeline requires interdisciplinary collaboration: AI researchers lead the development of predictive modeling, generative scenario synthesis, and causal reasoning techniques; automotive and transportation engineers contribute domain knowledge; human factors specialists incorporate realistic cognitive processes; and policymakers ensure that interventions are practical, reliable, and equitable.

If pursued, this agenda would transform traffic safety from a reactive forensics into a proactive prevention. By exposing emerging technologies to millions of synthetic yet plausible crash scenarios before deployment, it enables early identification of vulnerabilities. Real-time risk alerts can inform driver or system intervention, while causal insights guide targeted infrastructure improvements. Such science is essential to any credible path toward Vision-Zero.

\bibliography{reference}

\newpage

\appendix
\section*{Appendix}

\textbf{Roadmap.} In Section~\ref{proof:fish_rarity}, we establish the fundamental limits of rare‐event crash probability estimation. Section~\ref{proof:fish_stochastic} quantifies the Fisher information loss arising from unobserved stochastic factors. Section~\ref{proof:aug} introduces our counterfactual near‐miss augmentation framework, demonstrating how it amplifies Fisher information and reduces estimator variance. Section~\ref{sec:traffic_datasets} summarizes existing traffic datasets, including naturalistic driving datasets, autonomous vehicle datasets, and traditional crash records. Section~\ref{sec:traffic_simulators} reviews existing traffic simulation platforms, including traditional traffic simulators, autonomous driving simulators, and co-simulation frameworks. It also highlights emerging platforms enhanced by LLMs.

\section{Limits of Rare Crash Probability Estimation}
\label{proof:fish_rarity}

Crash occurrence \(Y_t\) is a binary outcome at each time step, modeled as $Y_t \sim \mathrm{Bernoulli}(p_t)$, where the instantaneous crash probability \(p_t\) lies between \(10^{-9}\) and \(10^{-6}\)~\cite{lord2005poisson,liu2024curse, yan2023learning}. Let  
\[
p = \mathbb{E}[p_t]
\]  
denote the average crash probability over time. For a Bernoulli model with parameter \(p\), the Fisher information, which quantifies how much the data informs the estimation of \(p\), is  
\[
I(p) = \frac{1}{p(1 - p)} \approx \frac{1}{p}, \quad \text{for } p \ll 1.
\]

The Cramér–Rao bound~\cite{kay1993fundamentals} gives a lower bound on the variance of any unbiased estimator \(\hat{p}\) of \(p\):
\[
\mathrm{Var}(\hat{p}) \ge \frac{1}{N I(p)} \approx \frac{p}{N},
\]
where \(N\) is the number of independent samples. The relative standard error (RSE) of \(\hat{p}\) is therefore
\[
\mathrm{RSE}
= \frac{\sqrt{\mathrm{Var}(\hat{p})}}{p}
\gtrsim \frac{1}{\sqrt{N p}}.
\label{eq:rse}
\]

This result highlights a fundamental challenge: when crash probabilities are extremely small, even extremely large datasets yield noisy estimates, making rare-event learning statistically inefficient.

\section{Fisher Information Loss from Unobserved Stochastic Factors}
\label{proof:fish_stochastic}

We define the full state of a traffic scene at time~\(t\) as \(Z_t = (X_t, E_t, H_t)\), where \(X_t\) represents observable vehicle kinematics, \(E_t\) denotes environmental conditions (e.g., weather, road surface), and \(H_t\) captures latent human states (e.g., attention, reaction time). While \(X_t\) is typically measurable, \(E_t\) and \(H_t\) evolve stochastically.

Because of the randomness in \(E_t\) and \(H_t\), the crash outcome \(Y_t\) remains uncertain even if \(X_t\) is fully known. This motivates defining the latent crash probability:
\[
p_t = \Pr(Y_t = 1 \mid X_t) = \mathbb{E}[Y_t \mid X_t].
\]
In practice, estimating the full conditional function \(p_t = \Pr(Y_t = 1 \mid X_t)\) is challenging, as environmental factors \(E_t\) and human states \(H_t\) are often only partially observed or measured with noise. This partial observability introduces additional variability into the relationship between \(X_t\) and \(Y_t\). As a result, most methods focus on estimating the average crash probability \(p = \mathbb{E}[p_t] = \Pr(Y_t = 1)\) using only the observed binary outcomes \(\{Y_t\}\).

The incomplete information about \(E_t\) and \(H_t\) weakens the statistical dependence between \(X_t\) and \(Y_t\) because it masks the underlying causal factors that contribute to crash risk. Mathematically, this loss of information arises from marginalizing over the unobserved factors:
\[
\Pr(Y_t = 1 \mid X_t) = \int \Pr(Y_t = 1 \mid X_t, E_t, H_t)\, p(E_t, H_t \mid X_t)\, dE_t\, dH_t.
\]
This marginalization smooths out variability in crash risk, reducing the curvature of the log-likelihood and thereby lowering the Fisher information. As a result, even large datasets offer limited precision in estimating \(p\), particularly when crashes are rare. 

\section{Augmentation via Counterfactual Near-Miss Generation}
\label{proof:aug}
To address the inefficiency of crash‐only estimation, we advocate a shift toward \emph{counterfactual augmentation}.  Many traffic samples do not result in crashes but exhibit high‐risk behaviors that correspond to elevated latent crash probability.  We therefore augment the dataset with \emph{near‐miss samples} satisfying
\[
\Pr(Y_t = 1 \mid Z_{t-\Delta:t}) \;>\;\tau,
\]
where \(Z_{t-\Delta:t}\) denotes traffic scene features over a short horizon.  Let
\[
\alpha
\;=\;
\Pr\bigl(\Pr(Y_t=1\mid Z_{t-\Delta:t})>\tau,\;Y_t=0\bigr)
\;\gg\;p,
\]
so that the augmented positive rate becomes
\[
p_{\mathrm{aug}}
\;=\; p + \alpha \gg p.
\]
Refer to Appendix~\ref{proof:fish_rarity}, the RSE improves from
\[
\text{RSE}(\hat p)
\;=\;\frac{1}{\sqrt{N\,p}}
\quad\longrightarrow\quad
\text{RSE}_{\mathrm{aug}}(\hat p)
\approx
\frac{1}{\sqrt{N\,(p+\alpha)}}
\;\ll\;
\frac{1}{\sqrt{N\,p}}.
\]
Thus, counterfactual near‐miss augmentation amplifies Fisher information and sharply reduces estimator variance \emph{without} waiting for additional observed crashes.

\section{Summary Table of Traffic Safety Datasets}
\label{sec:traffic_datasets}
This section reviews the principal data sources underpinning existing traffic safety research, broadly classified into three categories: traditional crash records (Table~\ref{tab:trad_crash_datasets}), naturalistic driving study (NDS, Table~\ref{tab:NDS}), and autonomous vehicle (AV) datasets (Table~\ref{tab:traffic_datasets}). Traditional crash record offer retrospective documentation of crash events but often lack detailed contextual or behavioral information. In contrast, NDS datasets provide continuous, real-world observations of driver behavior through instrumented vehicles equipped with various onboard sensors, enabling the analysis of pre-crash dynamics and near-miss incidents. Note that the NDS datasets in this study focus specifically on safety research. AV datasets represent a more recent advancement, further providing high-resolution, multi-modal sensor data collected from self-driving platforms.
\newcolumntype{L}[1]{>{\raggedright\arraybackslash}m{#1}} 
\newcolumntype{C}[1]{>{\centering\arraybackslash}m{#1}}  

\newcommand{\cmark}{\ding{51}} 
\newcommand{\diamondbullet}{\ding{117}}

\begin{table}[htbp]
  \caption{Overview of traditional crash record datasets}
  \scriptsize
  \centering
  \begin{threeparttable}
    \label{tab:trad_crash_datasets}
    \begin{tabularx}{0.9\linewidth}{
      @{} L{1.0cm} L{1.0cm} C{1.0cm} C{1.0cm} C{0.8cm} X @{}
    }
      \toprule
      \textbf{Dataset} & \textbf{Agency} & \textbf{Region}
        & \textbf{Sources\tnote{a}}
        & \textbf{Focus\tnote{b}}
        & \textbf{Primary use} \\
      \midrule
      CRIS   & TxDOT & TX       & $\square$            & $\oplus$                             & TX safety, trend analysis, identify high-risk corridors \\
      FHSMV  & FDOT  & FL       & $\square$            & $\oplus$                             & FL crash trend analysis, risk mapping \\
      NCDOT  & NCDOT & NC       & $\square$            & $\oplus$                             & NC policy planning \\
      SWITRS & CHP   & CA       & $\square$            & $\oplus$                             & CA crash analysis \\
      FARS   & NHTSA & US       & $\triangledown$      & $\oslash$                                   & National fatality trends \\
      CRSS   & NHTSA & US       & $\vartriangle$       & $\oplus$                             & Risk factor estimation \\
      CIREN  & NHTSA & US       & $\vartriangle$       & $\odot$                & Crash injury mechanisms, crash biomechanics \\
      HSIS   & FHWA  & States*  & $\lozenge$           & $\otimes$ & Design evaluation, SPF, CMF development \\
      \bottomrule
    \end{tabularx}

    \begin{tablenotes}[flushleft]\scriptsize
      \item \textbf{CRIS}: Crash records information system.
      \item \textbf{FHSMV}: Florida Highway Safety and Motor Vehicles.
      \item \textbf{NCDOT}: North Carolina Department of Transportation.
      \item \textbf{SWITRS}: Statewide Integrated Traffic Records System.
      \item \textbf{FARS}: Fatality Analysis Reporting System.
      \item \textbf{CRSS}: Crash Report Sampling System.
      \item \textbf{CIREN}: Crash Injury Research \& Engineering Network.
      \item \textbf{HSIS}: Highway Safety Information System.

      \item[a] \textbf{Sources}: [$\square$] Law enforcement; [$\vartriangle$] Police reports; [$\triangledown$] State safety agencies; [$\lozenge$] State DOTs.
      \item[b] \textbf{Focus}: [$\oplus$] Crashes result in property damage, injury, or death; [$\oslash$] Fatal; [$\odot$] Severe occupant injuries; [$\otimes$] Comprehensive database covering motor vehicle crashes, roadway inventory, traffic volumes, and more.
      \item[] \textbf{States*}: CA, IL, ME, MN, NC, WA, OH, and NC (Charlotte).
    \end{tablenotes}
  \end{threeparttable}
\end{table}

\newcommand{\dynMulti}{\ding{51}$^{\ast}$}

\begin{table}[htbp]
  \centering
  \scriptsize
  \caption{Overview of NDS dataset for safety study}
  \label{tab:NDS}
  \begin{threeparttable}
    \begin{tabularx}{\linewidth}{
      @{} 
      L{2.5cm}  
      L{1.8cm}  
      C{1.0cm}  
      L{1.5cm}  
      C{1.0cm}  
      C{1.2cm}  
      L{2.5cm}  
      @{}
    }
      \toprule
      \textbf{Dataset} &
      \textbf{Vehicle Class} &
      \textbf{\# Drivers} &
      \textbf{Timeline} &
      \textbf{VMT}\tnote{a} &
      \textbf{Region} &
      \textbf{Study Location} \\
      \midrule
      SHRP 2 \cite{Shrp2}        
        & Cars, SUVs, Minivans, Pickups   
        & 3100 
        & 36 months (2010–2013)  
        & 33 M  
        & US       
        & 6 cities (WA, NY, PA, NC, IN, FL) \\

      100 Car–NDS \cite{phase2006100} 
        & Cars, SUVs, Minivans            
        & 241  
        & 12–13 months (2004–2005) 
        & 2 M   
        & US       
        & 2 cities (Washington, Virginia) \\

      CNDS \cite{CNDS}               
        & Cars, SUVs, Minivans, Trucks    
        & 149  
        & 12–18 months (2012–2017)
        & 1.43 M
        & Canada   
        & 1 city (Saskatoon) \\

      ANDS \cite{ANDS}               
        & Cars, SUVs                       
        & 360  
        & 4 months (2015–2017)   
        & 0.93 M
        & Australia
        & 2 cities (Victoria, New South Wales) \\

      SH–NDS \cite{SHNDS}            
        & Cars, SUVs, Minivans             
        & 60   
        & 36 months (2012–2015)  
        & 0.1 M 
        & China    
        & 1 city (Shanghai) \\

      UDRIVE \cite{UDrive}           
        & Cars, truck, motorcycle          
        & 192  
        & 21 months (2012–2017)  
        & 1.43 M
        & Europe   
        & 6 countries (France, UK, Spain, Poland, Germany, Netherlands) \\
      \bottomrule
    \end{tabularx}

    \begin{tablenotes}[flushleft]\scriptsize
      \item \textbf{SHRP 2}: Second Strategic Highway Research Program.  
      \item \textbf{100car–NDS}: 100-Car Naturalistic Driving Study.
      \item \textbf{CNDS}: Canadian Naturalistic Driving Study.  
      \item \textbf{ANDS}: Australian Naturalistic Driving Study.  
      \item \textbf{SH-NDS}: Shanghai Naturalistic Driving Study.  
      \item \textbf{UDRIVE}: The European Naturalistic Driving Study.
      \item[a] \textbf{VMT}:Vehicle Miles Traveled refers to the total number of miles traveled by vehicles.
    \end{tablenotes}

  \end{threeparttable}
\end{table}

\renewcommand{\tabularxcolumn}[1]{m{#1}}
\newcolumntype{Z}{>{\centering\arraybackslash}m{1cm}}
\newcommand{\xmark}{\ding{55}}

\newcommand{\redcircle}{\tikz[baseline=-0.5ex]\node[draw=red, fill=red!20, text=black, circle, inner sep=0.4pt, minimum size=1em] {\textbf{C}};}
\newcommand{\bluecircle}{\tikz[baseline=-0.5ex]\node[draw=blue, fill=blue!20, text=black, circle, inner sep=0.4pt, minimum size=1em] {\textbf{L}};}

\newcommand{\orangecircle}{%
\tikz[baseline=-0.5ex] \node[draw=orange, fill=orange!20, text=black, circle, inner sep=0.4pt, minimum size=1em] {\textbf{R}};}

\newtcbox{\redbox}{on line, colback=red!20, colframe=red!80!black, boxrule=0.4pt, arc=0.5pt, boxsep=0pt, left=0.5pt, right=0.5pt, top=0.5pt, bottom=0.5pt}
\newtcbox{\bluebox}{on line, colback=blue!15, colframe=blue!80!black, boxrule=0.4pt, arc=0.5pt, boxsep=0pt, left=0.5pt, right=0.5pt, top=0.5pt, bottom=0.5pt}
\newtcbox{\yellowbox}{on line, colback=yellow!30, colframe=orange!90!black, boxrule=0.4pt, arc=0.5pt, boxsep=0pt, left=0.5pt, right=0.5pt, top=0.5pt, bottom=0.5pt}
\newtcbox{\greenbox}{on line, colback=green!20, colframe=green!70!black, boxrule=0.4pt, arc=0.5pt, boxsep=0pt, left=0.5pt, right=0.5pt, top=0.5pt, bottom=0.5pt}


\begin{table}[htbp]
\caption{Overview of representative open-source AV datasets from real-world and simulated environments}
\scriptsize
\centering
\begin{threeparttable}
\label{tab:traffic_datasets}
\begin{tabularx}{\linewidth}{@{}lZZZZZZX@{}}
\toprule
\textbf{Dataset} & \textbf{Crash}\tnote{a} & \textbf{Sensor}\tnote{b} & \textbf{Tasks}\tnote{c} & \textbf{Hz}\tnote{d} & 
\textbf{V2X}\tnote{e}& 
\textbf{HD Map}\tnote{f} & 
\textbf{Scope / Notes} \\
\midrule

\multicolumn{8}{l}{\textbf{Real-World Datasets}}\\ 
\midrule
nuScenes \cite{caesar2020nuscenes} & \xmark & \redcircle{} \bluecircle{} \orangecircle{}& \redbox{D} \bluebox{T} \yellowbox{S} \greenbox{M} & 2 & \xmark &\xmark& 40k urban AV scenes, 1000 segments, 23 class, 2 cities \\
Waymo \cite{ettinger2021large}           & \xmark & \redcircle{} \bluecircle{} & \redbox{D} \bluebox{T} \greenbox{M}  & 10  & \xmark &\cmark& 390k scenarios,1950 segments, 4 class, 6 cities, 12.6M trajectories \\
Argoverse-2 \cite{wilson2023argoverse}      & \xmark & \redcircle{} \bluecircle{} & \redbox{D} \bluebox{T} \yellowbox{S} \greenbox{M} & 10  & \xmark &\cmark& 250k scenarios, 113 segments, 11s duration, 10 class, 6 cities \\
Lyft Level 5 \cite{lyft2019}    & \xmark & \redcircle{} \bluecircle{} \orangecircle{} & \greenbox{M} & 10  & \xmark &\cmark& 170k urban AV instances, 25s duration, 10 class, 1 city \\
KITTI \cite{kitti}         & \xmark & \bluecircle{} & \redbox{D} \bluebox{T} \greenbox{M} & 10   & \xmark && 15k scenes, 10 segments, full-stack AV dataset \\
DAIR-V2X \cite{Dair-v2x}         & \xmark & \bluecircle{} & \redbox{D} & -- & V2I &\cmark& First real-world AV dataset comprise of 71254 frames of images  \\
V2X-Seq \cite{V2x-seq}        & \xmark & \bluecircle{} & \redbox{D} \bluebox{T} \greenbox{M} & 5-10  & V2I &\cmark& 15000 frames, 95 scenarios, 80,000 V2I scenarios, 672 driving hrs \\
YoutubeCrash \cite{kim2019crash}   & \cmark & \xmark & \xmark & --  & \xmark &\xmark& Public video crashes\\
TAD  \cite{tad}            & \cmark & \xmark & \xmark & --  & \xmark &\xmark& Surveillance crash footage only \\

\midrule
\multicolumn{8}{l}{\textbf{Simulator-Based Datasets}} \\
\midrule
OPV2V \cite{Opv2v}           & \xmark & \redcircle{} \bluecircle{} & \redbox{D} \bluebox{T} \greenbox{M} & 10  & V2V &\xmark& 33k multi-agent V2V sim, 18k V2X frames, 8 digital towns in CARLA \\
V2X-Sim \cite{V2X-Sim}         & \xmark & \redcircle{} \bluecircle{} & \redbox{D} \bluebox{T} \yellowbox{S} \greenbox{M} & 5   & V2V, V2I &\xmark& 47k of annotated samples , 10k V2X co-sim scenarios \\
VIENA$^2$ \cite{VIENA2}       & \cmark & \xmark & \xmark & --  & \xmark &\xmark& Synthetic rare-event videos, crash types \\
GTACrash \cite{kim2019crash}         & \cmark & \xmark & \xmark & --  & \xmark &\xmark& Simulated crash clips \\
DeepAccident  \cite{deepaccident}   & \cmark & \redcircle{} \bluecircle{} \orangecircle{}& \redbox{D} \bluebox{T} \yellowbox{S} \greenbox{M}& 10  & V2V, V2I &\cmark& 285k samples, 57k V2X, proactive safety sim \\
Shift  \cite{shift}   &\xmark &\redcircle{} \bluecircle{} & \yellowbox{S} \greenbox{M}& 10 & \xmark &\xmark& 4850 segments, 33min duration, 2.5M frames, captured 8 cities  \\
\bottomrule
\end{tabularx}

\begin{tablenotes}[flushleft]
\scriptsize
\item[a] \textbf{Crash}: Labeled crash or near-miss events are present.
\item[b] \textbf{Sensor}: Featuring high-resolution sensors Camera: \redcircle{}, LiDAR: \bluecircle{}, Radar: \orangecircle{}.
\item[c] \textbf{Tasks}: Supports tasks such as Detection: \redbox{\scriptsize D}, Tracking: \bluebox{\scriptsize T}, Segmentation: \yellowbox{\scriptsize S}, Motion Forecasting: \greenbox{\scriptsize M}.
\item[d] \textbf{Hz}: Approximate sampling frequency (Hz); "--" indicates not reported.
\item[e] \textbf{V2X}: Scenario constructed Vehicle-to-Vehicle (V2V), Vehicle-to-Infrastructure (V2I), for None :\xmark{}.
\end{tablenotes}

\end{threeparttable}
\end{table}


\newpage
\section{Summary Table of Traffic Simulation Platforms}
\label{sec:traffic_simulators}
In this section, we summarize existing traffic simulators (Table~\ref{tab:simtool_overview}), including traditional traffic simulation platforms, autonomous driving simulators, and co-simulation frameworks. Emerging simulation platforms enhanced by LLMs are also included (Table~\ref{tab:llm_simtool_ow}). The evaluation focuses on simulator characteristics relevant to traffic safety analysis. Specifically, features such as 3D environment modeling and high-fidelity vehicle dynamics are emphasized, as they are essential for safety-critical scenario analysis. In contrast, features less directly related to safety, such as V2X or V2V communication capabilities, are not considered in this review.  

\newcolumntype{Y}{>{\centering\arraybackslash}m{1.4cm}}
\begin{table}[!h]
\caption{Capabilities of major traffic simulation platforms}
\scriptsize
\centering
\begin{threeparttable}
\label{tab:simtool_overview}
\begin{tabularx}{\textwidth}{@{}p{3.0cm}YYYYYY@{}}
\toprule
\textbf{Name} &
\textbf{CityTF}\tnote{a} &
\textbf{3DEnv}\tnote{b} &
\textbf{Dyn}\tnote{c} &
\textbf{Sensors}\tnote{d} &
\textbf{VehInt}\tnote{e} &
\textbf{Open\tnote{f}}
\\
\midrule
\multicolumn{7}{l}{\textbf{Traffic Microscopic / Mesoscopic Simulators}}\\
\midrule
SUMO \cite{krajzewicz2002sumo}                & \cmark & \xmark & \xmark & \xmark & \dynMulti & \cmark\\
PTV Vissim \cite{PTV_Vissim_2025}             & \cmark & \dynMulti & \xmark & \xmark & \dynMulti & \xmark\\
Aimsun Next \cite{Aimsun_Next_24}             & \cmark & \dynMulti & \xmark & \xmark & \dynMulti &\xmark\\
Paramics Discovery \cite{Paramics_Discovery_2025} & \cmark & \dynMulti & \xmark & \xmark & \dynMulti & \xmark\\
CORSIM \cite{halati1997corsim}                & \cmark & \xmark & \xmark & \xmark & \dynMulti & \xmark\\
\midrule
\multicolumn{7}{l}{\textbf{High-Fidelity Vehicle / Driving \& AV Simulators}}\\
\midrule
CARLA \cite{dosovitskiy2017carla}             & \xmark & \cmark & \dynMulti & \cmark & \dynMulti & \cmark\\
Simcenter PreScan \cite{Siemens_Prescan_2503} & \xmark & \cmark & \dynMulti & \cmark & \dynMulti & \xmark\\
IPG CarMaker \cite{IPG_CarMaker_14}           & \xmark & \cmark & \dynMulti & \cmark & \dynMulti & \xmark\\
VIRES VTD \cite{VTD_25}                       & \cmark & \cmark & \dynMulti & \cmark & \dynMulti & \xmark\\
LG SVL Simulator \cite{rong2020lgsvl}         & \xmark & \cmark & \dynMulti & \cmark & \dynMulti & \cmark\\
Gazebo                \cite{koenig2004design} & \xmark & \cmark & \dynMulti & \cmark & \xmark & \cmark\\
Project Chrono \cite{serban2018chrono}        & \xmark & \cmark & \cmark & \cmark & \xmark & \cmark\\
NVIDIA PhysX \cite{NVIDIA_PhysX_5_6_0}        & \xmark & \xmark & \dynMulti & \xmark & \xmark & \cmark\\
SynChrono \cite{taves2020synchrono}           & \xmark & \cmark & \cmark & \cmark & \dynMulti & \cmark\\
\midrule
\multicolumn{7}{l}{\textbf{Co-Simulation Frameworks}}\\
\midrule
CARLA+SUMO+PhysX \cite{zhang2025virtual}  & \cmark & \cmark & \dynMulti & \cmark  & \dynMulti & -\\
CARLA+Vissim \cite{erdagi2025development}   & \cmark & \cmark & \dynMulti & \cmark & \dynMulti & -\\
PreScan+Vissim \cite{chen2024safety}        & \cmark & \cmark & \dynMulti & \cmark & \dynMulti & -\\
SUMO+OMNeT++ \cite{shuai2022co}       & \cmark & \xmark & \xmark & \xmark & \dynMulti &  -\\
\bottomrule
\end{tabularx}
\begin{tablenotes}[flushleft]\footnotesize
\scriptsize
\item[a] \textbf{CityTF}: traffic flow with rule-based,  car-following, lane-changing, and signal control.
\item[b] \textbf{3DEnv}: 3-D road geometry, buildings, weather effects (\dynMulti = Presentation-grade, \cmark = Sensor/physics-grade).
\item[c] \textbf{Dyn}: high-fidelity vehicle dynamics level (\dynMulti = multi-body, \cmark = high-order multi-body + deformable-terrain).
\item[d] \textbf{Sensors}: native virtual sensors (camera, LiDAR, radar, etc.).
\item[e] \textbf{VehInt}: interaction of multiple road-user classes (vehicle, pedestrian, etc.) (\dynMulti = pair-wise interaction, \cmark = group-wise interaction)~\cite{wu2025ai2}.
\item[f] \textbf{Open}: open source or commercial.
\end{tablenotes}
\end{threeparttable}
\end{table}

\begin{table}[!h]
    \caption{Key capabilities of LLM-enhanced simulation frameworks}
    \scriptsize
    \centering
    \begin{threeparttable}
        \label{tab:llm_simtool_ow}
        \begin{tabularx}{\linewidth}{@{}p{2.7cm}YYYYYYY@{}}
        \toprule
        \textbf{Framework} &
        \textbf{SceneGen}\tnote{a} &
        \textbf{PolicyGen}\tnote{b} &
        \textbf{Coverage}\tnote{c} &
        \textbf{WorldBuild}\tnote{d} &
        \textbf{VehInt}\tnote{e} &
        \textbf{Multimodal}\tnote{f} \\ 
        \midrule
        ChatScene+CARLA \cite{zhang2024chatscene}            & \cmark & \xmark & \cmark & \xmark & \dynMulti & \xmark  \\
        CodeLLM+CARLA \cite{ma2024learning}        & \xmark & \cmark & \xmark & \xmark & \xmark & \xmark \\
        ChatSUMO \cite{li2024chatsumo}                  & \cmark & \xmark & \xmark & \xmark & \xmark & \xmark \\
        ChatGPT+Omniverse\cite{NVIDIA_AIRoomGen_2025}         & \cmark & \xmark & \xmark & \cmark & \xmark & \xmark \\
        ChatGPT+CARLA \cite{ruan2024ttsg}                     & \cmark & \xmark & \cmark & \xmark & \dynMulti & \xmark \\
        \bottomrule
        \end{tabularx}
        \begin{tablenotes}[flushleft]\footnotesize
            \scriptsize
            \item[a] \textbf{SceneGen}: text-to-scenario generation (roads, actors, states).
            \item[b] \textbf{PolicyGen}: LLM writes or edits driving-policy code.
            \item[c] \textbf{Coverage}: generate both ordinary (day-to-day) traffic scenes and safety-critical  scenarios.
            \item[d] \textbf{WorldBuild}: 3-D asset / map generation or placement (e.g., weather conditions, 3D road geometry).
            \item[e] \textbf{VehInt}: interaction of multiple road-user classes (vehicle, pedestrian, etc.) (\dynMulti = pair-wise interaction, \cmark = group-wise interaction)~\cite{wu2025ai2}.
            \item[f] \textbf{Multimodal}: multimodal human drivers.
        \end{tablenotes}
    \end{threeparttable}
\end{table}

\end{document}